\documentclass[lettersize,journal]{IEEEtran}  

\IEEEoverridecommandlockouts                              

\usepackage{tabularx}
\usepackage{booktabs}%
\usepackage{amsmath} 
\usepackage{graphicx}
\usepackage{xcolor}
\usepackage{comment}

\usepackage{balance}

\title{
Robot-Led Vision Language Model Wellbeing Assessment of Children
}

\author{Nida Itrat Abbasi$^{1,\dagger,*}$, \and Fethiye Irmak Dogan$^{1,\dagger}$, \and Guy Laban$^{1,\dagger}$, \and Joanna Anderson$^{2}$, \\ \and Tamsin Ford$^{2}$, \and Peter B. Jones$^{2}$, \and Hatice Gunes$^{1}$\\
{\normalsize
    $^1$ Department of Computer Science and Technology, University of Cambridge, Cambridge, United Kingdom\\
    $^2$ Department of Psychiatry, University of Cambridge, Cambridge, United Kingdom}
\thanks{$\dagger$ Equal Contribution, * Corresponding author}
\thanks{N. I. Abbasi is supported by the W.D. Armstrong Trust PhD Studentship and the Cambridge Trusts. F. I. Dogan, G. Laban, and H. Gunes have been supported by the EPSRC project ARoEQ under grant ref. EP/R030782/1. All research at the Department of Psychiatry (University of Cambridge) is supported by the NIHR Cambridge Biomedical Research Centre (BRC-1215-20014, particularly T. Ford) and NIHR Applied Research Collaboration East of England (P. Jones, J. Anderson). The views expressed are those of the authors and not necessarily those of the NIHR or the Department of Health and Social Care. \textbf{Open Access:} For the purpose of open access, the authors have applied a Creative Commons Attribution (CC BY) licence to any Author Accepted Manuscript version arising. \textbf{Data Access:} Overall statistical analysis of research data underpinning this publication is available in the text of this publication. Additional raw data related to this publication cannot be openly released; the raw data contains transcripts of interviews, but none of the interviewees consented to data sharing.}}%

\begin{document}

\maketitle
\thispagestyle{empty}
\pagestyle{empty}

\begin{abstract}

This study presents a novel robot-led approach to assessing children’s mental wellbeing using a Vision Language Model (VLM). Inspired by the Child Apperception Test (CAT), the social robot NAO presented children with pictorial stimuli to elicit their verbal narratives of the images, which were then evaluated by a VLM in accordance with CAT assessment guidelines. The VLM’s assessments were systematically compared to those provided by a trained psychologist. The results reveal that while the VLM demonstrates moderate reliability in identifying cases with no wellbeing concerns, its ability to accurately classify assessments with clinical concern remains limited. Moreover, although the model’s performance was generally consistent when prompted with varying demographic factors such as age and gender, a significantly higher false positive rate was observed for girls, indicating potential sensitivity to gender attribute. These findings highlight both the promise and the challenges of integrating VLMs into robot-led assessments of children's wellbeing.

\end{abstract}

\section{Introduction}
The rising prevalence of mental health and well-being issues among children \cite{Lachaab2024Trends2016-2022} emphasises the need to develop solutions that are engaging, reliable, accurate and accessible to all. 
One potential direction is the use of social robots~\cite{Kabacinska2021SociallyReview,2022SocialTreatment}, which have been shown to encourage self-disclosure~\cite{Abbasi2022CanStudy,Abbasi2025ARobot,laban2024building,ced_2023}, thereby encouraging individuals to express their emotions and feelings with minimal social repercussions~\cite{2024Sharing}. Considering the ongoing introduction of conversational AI techniques, such as Large Language Models (LLMs) and Vision Language Models (VLMs), into social robots~\cite{10715872, rahimi2025user}, these agents can communicate better while also using their capabilities to reason and make inferences about users. Therefore, beyond enhancing social interactions and facilitating potential mental health treatment and interventions~\cite{SpitaleMicol2025VITA:Coaching,2025AReappraisal}, the integration of LLMs and VLMs could also support the facilitation of wellbeing assessments. 


In this study, we aim to evaluate the accuracy, consistency, and potential sensitivities of a robot-led VLM wellbeing assessment of children. First, due to the imperative to confirm the measurement precision~\cite{Saal1980RatingData,Abbasi2024RobotisingInteractions}, 
of such novel approach, we systematically compare its outcomes with those of trained psychologists to determine its accuracy in capturing children's wellbeing. Therefore, we are asking: \textit{\textbf{(RQ1)} To what extent do robot-led VLM assessments of children's wellbeing align with those of a psychologist?} Moreover, LLMs and VLMs tend to adapt and generalize outputs to varying input details
~\cite{radford2019language}, which results in sensitivity to prompts wording and key features \cite{Zhao2021CalibrateModels,10.1145/3411763.3451760}. 
Therefore, it is critical to assess whether a robot-led VLM wellbeing assessment maintains consistent assessments based on the core wellbeing indicators despite variations in demographic attributes such as gender and age. Accordingly, we also ask: \textit{\textbf{(RQ2)} To what extent are robot-led VLM assessments of children's well-being consistent when prompted with varying demographic information, specifically gender and age?} Finally, automated assessments of children’s wellbeing, much like human evaluations \cite{Snowden2003BiasEvidence}, can be influenced by a range of socio-demographic attributes (e.g., \cite{Omar2024Socio-DemographicAnalysis}). Since LLMs and VLMs are trained on human-generated data, there is a 
concern that these models may perpetuate or even exacerbate existing healthcare biases, which often reflect societal prejudices present in the training data \cite{Guo2024BiasMitigation,Taubenfeld2024SystematicDebates}. Gender, in particular, has been shown to systematically influence the interpretation and treatment of patient data in human assessments ~\cite{Hertler2020Sex-specificOutcomes} and subsequently in AI-driven systems ~\cite{cheong2023s, cheong2024small, spitale2024appropriateness}, often due to ingrained stereotypes or implicit biases ~\cite{Garb2021RaceDisorders}. Accordingly, it is crucial to examine whether robot-led VLM assessments encode implicit sensitivity to critical attributes such as gender. Thus, 
we ask: \textit{\textbf{(RQ3)} To what extent are robot-led VLM assessments of children’s wellbeing sensitive to the gender attribute?}

We conducted a secondary analysis to~\cite{Abbasi2022CanStudy, abbasi2024analysing}, in which 36 children participated in a Child Apperception Test (CAT)~\cite{bellak1949children} conducted by the NAO robot\footnote{https://aldebaran.com/en/support/kb/category/nao6/} to assess their well-being. The children's verbal narrations in response to the CAT's pictorial stimuli were analysed by VLM following the CAT guidelines~\cite{bellak1949children}, using prompts addressing the child's demographic information (age and gender). The VLM's assessments were then compared with those provided by a trained psychologist who employed the same criteria. It is essential to note that this work does not seek to replace psychologists but to examine whether and how VLMs might augment their decision-making in robot-led assessments of children’s wellbeing. 

\section{Related Works}

Social robots effectively conducted standardised psychological assessments 
in children in both in-person~\cite{Abbasi2022CanStudy} and online settings~\cite{Abbasi2025ARobot} and were found successful in estimating children's mood for behavioural adaptation and providing them with emotional support~\cite{gamborino2019mood}. In addition, robots have also been preferred by children to share sensitive information like bullying experiences over human counterparts~\cite{bethel2016using}. As such, robots are often considered to have the potential to help children open up and articulate complex emotions by appearing non-intimidating and engaging, which helps them feel at ease~\cite{Marchetti2022RoboticsPsychology}. 

Previous studies also show that LLMs can enhance user interaction experiences in HRIs by making the robot's communication capabilities more intelligent, personalised, and context-aware~\cite{spitale2024appropriateness,2025AReappraisal}. However, when temporal and multimodal information are involved, further investigation is needed to understand real-world interactions~\cite{Hu2025SimulatingModels}. Therefore, while VLMs are not yet widely used to facilitate social interactions in HRI~\cite{Pashangpour2024TheHealthcare}, their potential—along with that of LLMs—to support robot reasoning, assessment, evaluation, and classification is increasingly being explored. For example, Dogan et al.~\cite{dogan2024grace} have explored how LLMs can be used to enhance the socially appropriate decision-making process of robots in human environments showing that LLMs predications can (and probably should) be improved by following human preferences and insights. Nonetheless, Abbo and Belpame~\cite{abbo2025vision} report that VLM outputs are not always aligned with human judgments when identifying value-laden elements in domestic scenarios, with even the best-performing model achieving low alignment scores; however, 
they suggest that with appropriate fine-tuning and prompt engineering, these models could potentially enhance 
contextual sensitivity in social robots. These results correspond well to Etesam et al.~\cite{10802538}, which demonstrated that VLMs, when fine-tuned even on small datasets, significantly outperform traditional baselines in contextual emotion recognition. 

Complementing these advances in multimodal perception, other works examined how LLMs internally represent and reason about affective states. Tak and Gratch~\cite{Tak2023IsEmotion} found that GPT-4 effectively predicts emotional appraisals and valence but struggles with emotional intensity and coping strategies, while their follow-up work~\cite{Tak2024GPT-4Perspective} showed that GPT-4 aligns more closely with third-person (observer) emotion attributions than with self-assessments, suggesting LLMs may adopt an average-human perspective that could be beneficial for modeling social perceptions in HRI. 
Mehra et al. ~\cite{mehra2025beyond} evaluated LLMs' ability to classify and explain facial expressions from valence and arousal values, finding that while LLMs performed well in semantically describing facial expressions, they were limited in classifying these into categories of emotions. Together, these findings highlight both the promise and the current limitations of integrating LLMs and VLMs into robot-led assessment tasks. While these models show growing capabilities in affective reasoning and multimodal understanding, careful consideration of their accuracy, consistency, and alignment with human interpretations remains essential for their responsible deployment.

It's important to consider that sensitive attributes such as gender, age, socioeconomic background, etc., have been shown to affect VLM/LLM outputs. For example, a previous study with about 500 emergency departments, assessing nine LLMs, found that certain marginalized groups, such as coloured unhoused patients, were more likely to be referred for urgent care procedures and mental health assessments, while the high-income population were more likely to be recommended for advanced diagnostics~\cite{Omar2024Socio-DemographicAnalysis}. Another example shows that in text generation tasks, GPT models produced more agentic descriptions for males and more communal descriptions for women~\cite{kaplan2024s}. Moreover, Visual linguistic models such as VL-BERT have been shown to at times favour gender stereotypes over visual evidence in their output generation \cite{srinivasan2021worst}. 
Hence, understanding these limitations is critical to developing ethical approaches that provide accurate assessments, especially in sensitive areas of the mental health of children.

\section{Methodology}


\subsection{Dataset}
\label{sec:dc}

This work uses a dataset described in \cite{Abbasi2022CanStudy, abbasi2024analysing} with 
36 children (\textit{M}age = 9.44, \textit{STD} = 1.36; 19 boys and 17 girls) 
who visited the lab and interacted with the robot in a dyadic setting (see Figure~\ref{pipeline}(a)). 
Informed consent was obtained from the participants' guardians. 
The robot facilitated a CAT~\cite{bellak1949children} inspired interaction where   
children were presented with pictorial stimuli (cards 7, 9, and 10, see Figure~\ref{pipeline}(b)) and were asked to explain them. 
A clinically-trained psychologist assessed the child's narrative following the CAT~\cite{bellak1949children} guidelines. Further details on the sections, items, and the scoring framework can be found 
in~\cite{bellak1949children}. 

\subsection{Procedure}
\label{sec:pro}

\begin{figure}
    \centering
    \includegraphics[width=0.95\linewidth]{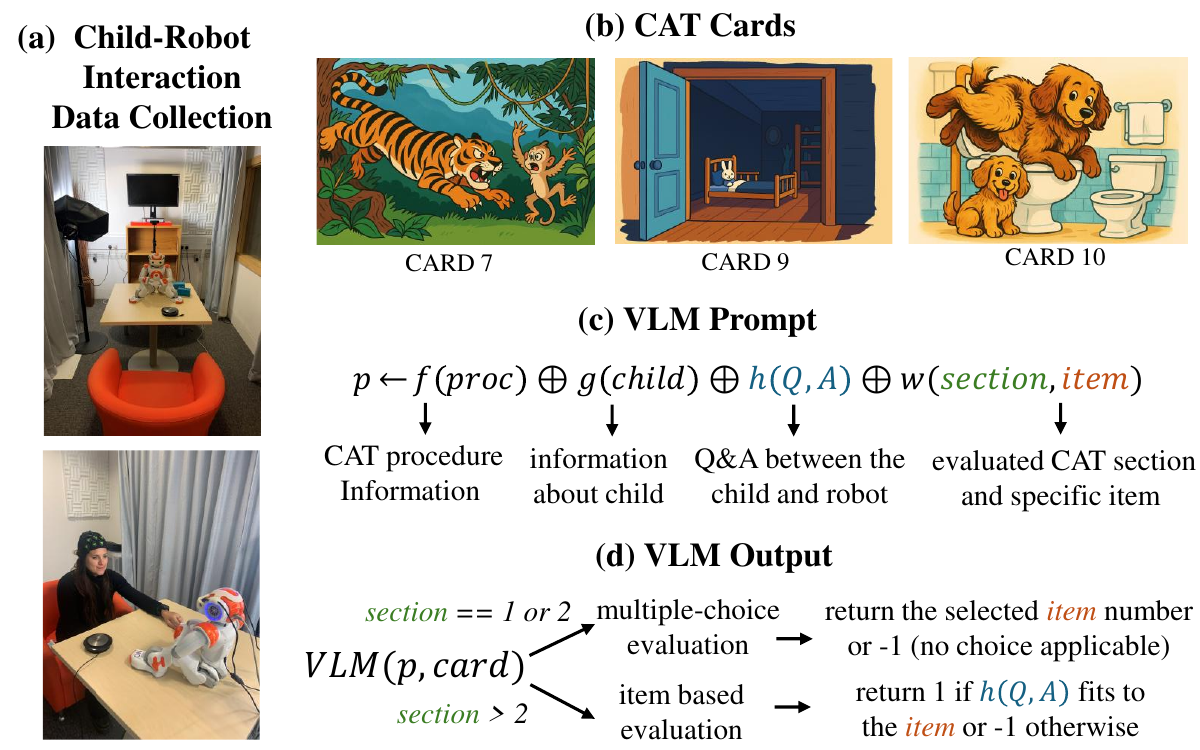}
    \caption{ (a) The CRI data collection setup, (b) the CAT cards~\cite{bellak1949children} shown to the children and provided to the VLM (cartoonized by OpenAI Sora for displaying in the paper), (c) the VLM prompt, and (d) the VLM output.}
    \label{pipeline}
    \vspace{-2em}
\end{figure}
First, the following prompt was constructed for facilitating the CAT~\cite{bellak1949children} assessment using VLMs: 
\begin{equation}
p \leftarrow f(proc) \oplus g(child) \oplus h(Q, A) \oplus w(section, item),
\end{equation}%
\noindent where $f(proc)$ represents the information about the CAT procedure (see \cite{bellak1949children}). 
Additionally, $g(child)$ denotes the string that includes information about the child (the child's gender, age, both or neither). $h(Q, A)$ contains the child's narration of the CAT's pictorial stimuli when responding to each of the robot's questions. 
$w(item, express)$ outputs a string for each CAT's section (e.g., `\textit{repression and denial}') combined with an assessment item (e.g., `\textit{Child omits figures or objects from story}') that the VLM was prompted to assess if it appears in the child's disclosed narrative. Further, $\oplus$ represents the string concatenation to form the VLM prompt $p$.  Finally, the VLM prompt 
forced the assessment to be returned as a numerical value   (i.e., `\textit{Return the most appropriate evaluation number, and nothing else. If none of the evaluations is appropriate, return -1}') (see Figure~\ref{pipeline}(c)).

Since the first two sections in the CAT manual consist of multiple-choice questions \cite{bellak1949children}, $w(section, item)$ included all assessment items at once, and the VLM was prompted to select an option that is applicable for provided $h(Q, A)$ (e.g., `Option 2') or return `-1' if none applied. In the remaining sections, following the CAT procedure, each item was assessed individually based on whether the child’s disclosure fit ('1') or did not fit ('-1') the item 
(see Figure~\ref{pipeline}(d)).

The VLM \textit{`llava-v1.6-mistral-7b-hf'}\footnote{https://huggingface.co/llava-hf/llava-v1.6-mistral-7b-hf} was chosen for the task for safety and ethical reasons as being a lightweight, open-source model that allows local processing of data. It was queried with three different CAT cards separately, $VLM(p, card)$. 
The VLM was restricted to generating a maximum of 20 tokens, and the output was decoded by skipping special tokens.
Finally, 
string outputs were converted to numerical values using a "bag-of-words" technique, replacing 
strings such as `does not fit' and `does not involve'. 

In total, the VLM was prompted to assess 27652 items on the Department of Computer Science and Technology of the University of Cambridge High-Performance Computing (HPC) system equipped with an AMD EPYC 7763 64-core processor, 1000 GiB RAM, and an NVIDIA A100-SXM-80GB GPU. The whole process took approximately 28 hours.
 \subsection{Statistical Analysis}
 \label{sec:sa}
We refer to the VLM's evaluation as \texttt{VLM}, the output generated when prompted with demographic information as \texttt{VLM (X)}  where \texttt{X} can be age \texttt{(A)}, gender \texttt{(G)} or 
both \texttt{(A+G)}, and the psychologist's assessment as \texttt{'Psychologist'}. 
A Cohen’s Kappa analysis was conducted to assess the agreement between the models and the psychologist in their assessments, used as a measure of inter-rater reliability~\cite{Saal1980RatingData}. 
A Bayesian Beta-Binomial analysis was conducted to evaluate the consistency between different assessment results of the models and the psychologist assessment as ground truth. 
The analysis utilized a Beta-Binomial model with uniform prior parameters ($\alpha = 1$, $\beta = 1$). The posterior distributions were estimated for each pairwise comparison by computing the number of matching outputs (successes) and the total number of trials. The analysis further examined agreement for both negative assessments (-1) and positive assessments (1) separately to evaluate the consistency of model-generated outputs in capturing the ground truth. We computed Bayesian 95\% Credible Intervals (CrIs) for the differences between model agreement rates. These CrIs estimate the extent to which one model differs from another, with positive values indicating higher agreement for the first model and negative values indicating higher agreement for the second.
An Equalized Odds (EO) test was conducted to assess whether the classifier’s error rates were equivalent across gender groups. According to the EO criterion, a classifier should yield comparable true positive rates (TPR) and false positive rates (FPR) across groups~\cite{NIPS2016_9d268236}. To statistically evaluate whether the observed differences were significant, two-proportion z-tests were performed~\cite{Diciccio2020}. 
In addition, we used Spearman's correlation to evaluate whether age as a sensitive attribute affected VLM's outputs.

\subsection{Pilot Evaluation}

We compared three VLM prompting strategies: (1) providing the VLM with the original CAT manual, 
(2) using a simplified version of the CAT manual, 
and (3) omitting the CAT manual 
for testing the model's foundational knowledge of the CAT exercise. We assessed the inter-rater reliability of each method by computing Cohen’s Kappa scores against the human psychologist assessments. Although the differences in Kappa scores were minimal across the conditions ($\kappa$ = 0.18-0.22), the model prompted with the original CAT manual ($\kappa$ = 0.22) achieved the highest agreement (fair agreement). Accordingly, it was the selected approach for the analysis.

\section{Results}

\subsection{Evaluating Agreement
}

Table \ref{kappa} presents the pairwise Cohen’s Kappa scores for agreement between the different models and the psychologists. The agreement between \texttt{VLM} and \texttt{VLM (G)} was $\kappa = 0.98$, indicating almost perfect agreement~\cite{Landis1977a,Landis1977b}. Similarly, \texttt{VLM (A)} showed high agreement with \texttt{VLM (A+G)} ($\kappa = 0.98$). Agreement between \texttt{VLM} and \texttt{VLM (A)} was also strong ($\kappa = 0.85$), while \texttt{VLM} and \texttt{VLM (A+G)} had a slightly lower but still substantial agreement ($\kappa = 0.84$).

In contrast, the agreement between any of the four VLM-produced assessments and the psychologist's assessments was substantially lower, ranging from $\kappa = 0.20$ (\texttt{VLM (A)}) to $\kappa = 0.22$ (\texttt{VLM} \& \texttt{VLM (G)}), indicating only a fair agreement between the models and the psychologist. These results suggest that the different prompting approaches were highly consistent with each other, but their alignment with the human psychologists was relatively low, as the human assessments diverged more significantly from the automated VLM outputs. 

The Bayesian Beta-Binomial analysis assessed the agreement between each output and the ground truth (\texttt{Psychologist}) (see Table \ref{tab:bes}). Overall, high agreement rates were observed across outputs when considering all cases, with \texttt{VLM} showing an estimated probability of agreement with \texttt{Psychologist} of 82.25\% (95\%CI [81.33\%, 83.15\%]). However, the results revealed significant variation when evaluating positive and negative classifications separately.

For negative assessments (-1), the agreement between \texttt{VLM} and \texttt{Psychologist} was 82.25\% (95\%CI [81.33\%, 83.15\%]), indicating moderate reliability in detecting cases where no assessment was made. \texttt{VLM (G)} exhibited a similar agreement probability of 81.53\% (95\%CI [80.60\%, 82.45\%]), while \texttt{VLM (A)} showed a slightly higher agreement of 82.66\% (95\%CI [81.75\%, 83.55\%]). \texttt{VLM (A+G)} had an agreement probability of 82.13\% (95\%CI [81.23\%, 83.03\%]), further confirming the overall trend.

For positive assessments (1), the agreement rates were substantially lower across all comparisons. The estimated probability of agreement between \texttt{VLM} and \texttt{Psychologist} for cases labelled as 1 was only 13.38\% (95\%CI [10.64\%, 16.38\%]), suggesting a high degree of inconsistency when identifying actual assessments. \texttt{VLM (G)} exhibited a nearly identical trend, with an agreement probability of 13.30\% (95\%CI [10.55\%, 16.30\%]). \texttt{VLM (A)} showed an agreement probability of 14.02\% (95\%CI [11.07\%, 17.44\%]), while \texttt{VLM (A+G)} demonstrated slightly higher agreement at 15.21\% (95\%CI [12.08\%, 18.74\%]). These findings indicate that while the models were moderately reliable at recognizing negative cases (i.e., cases with no assessment for low wellbeing), they demonstrate substantial difficulty in correctly identifying actual assessments (i.e., cases assessed with higher prevalence for low wellbeing).

\begin{table}[h!]
    \centering
    \caption{Cohen's Kappa Agreement Scores.
    }\label{kappa}
    \scriptsize
    \renewcommand{\arraystretch}{1} %
    \begin{tabularx}
    {0.49\textwidth}{lXXXXX}
    \toprule
    {} &  \texttt{VLM} & \texttt{VLM (G)} &  \texttt{VLM (A)} &  \texttt{VLM  (A+G)} &  \texttt{Psych.} \\
    \midrule
    \texttt{VLM}                 &              1.00 &                      0.98 &                   0.85 &                              0.84 &          0.22 \\
    \texttt{VLM (G)}         &              0.98 &                      1.00 &                   0.85 &                              0.84 &          0.22 \\
    \texttt{VLM (A)}           &              0.85 &                      0.85 &                   1.00 &                              0.98 &          0.20 \\
    \texttt{VLM (A + G)} &              0.84 &                      0.84 &                   0.98 &                              1.00 &          0.20 \\
    \texttt{Psychologist}                     &              0.22 &                      0.22 &                   0.20 &                              0.20 &          1.00 \\
    \bottomrule
    \end{tabularx}
    \vspace{-1em}
\end{table}

\subsection{Evaluating Consistency
}

To further explore inconsistencies in assessment across the VLMs (\texttt{VLM}, \texttt{VLM (G)}, \texttt{VLM (A)}, \texttt{VLM (A + G)}), an additional analysis was conducted to examine mismatches where VLM outputs were -1 while \texttt{Psychologist} was 1 and vice versa. The results showed that when \texttt{VLM} classified an instance as -1 while the ground truth was 1, the probability of such mismatches occurring was 3.19\% (95\%CI [2.76\%, 3.65\%]). Conversely, when \texttt{VLM} classified an instance as 1 while the ground truth was -1, the mismatch probability was 3.46\% (95\%CI [3.03\%, 3.91\%]). Similar trends were observed across the other parsing conditions, with mismatch probabilities ranging from 3.03\% to 3.40\%. These findings indicate that while overall agreement is moderate, there are systematic errors in classification where certain cases are frequently misidentified in both directions.

\begin{table}[h!]
\centering
\caption{
Bayesian Beta-Binomial Analysis Results. 
}\label{tab:bes}\resizebox{\columnwidth}{!}{
\begin{tabular}{lcccc}
\toprule
\textbf{ } & \textbf{Successes} & \textbf{Total Cases} & \textbf{Mean Probability} & \textbf{\textit{95\%CI}} \\
\midrule
\multicolumn{3}{c} {\scriptsize\textbf{No-Assessment}
(Model and Psychologist output = -1)} & &\\
\texttt{VLM} & 5588 & 6793 & 82.25\% & [81.33\%, 83.15\%] \\
\texttt{VLM (G)} & 5535 & 6788 & 81.53\% & [80.60\%, 82.45\%] \\
\texttt{VLM (A)} & 5608 & 6784 & 82.66\% & [81.75\%, 83.55\%] \\
\texttt{VLM (A+G)} & 5567 & 6780 & 82.13\% & [81.23\%, 83.03\%] \\
\midrule
\multicolumn{3}{c} {\scriptsize\textbf{Assessment} (Model and Psychologist output = 1)} & &\\
\texttt{VLM} & 71 & 536 & 13.38\% & [10.64\%, 16.38\%] \\
\texttt{VLM (G)} & 70 & 532 & 13.30\% & [10.55\%, 16.30\%] \\
\texttt{VLM (A)} & 75 & 535 & 14.02\% & [11.07\%, 17.44\%] \\
\texttt{VLM (A+G)} & 81 & 533 & 15.21\% & [12.08\%, 18.74\%] \\
\midrule
\multicolumn{3}{c} { \scriptsize
\textbf{False Negative} (Model output: -1, Psychologist = 1)} & & \\
\texttt{VLM} & 187 & 5894 & 3.19\% & [2.76\%, 3.65\%] \\
\texttt{VLM (G)} & 182 & 5841 & 3.13\% & [2.70\%, 3.59\%] \\
\texttt{VLM (A)} & 178 & 5914 & 3.03\% & [2.60\%, 3.48\%] \\
\texttt{VLM (A+G)} & 183 & 5886 & 3.11\% & [2.68\%, 3.56\%] \\
\midrule
\multicolumn{3}{c} {\scriptsize
\textbf{False Positive} (Model Output = 1, Psychologist = -1)} & &\\
\texttt{VLM} & 230 & 6677 & 3.46\% & [3.03\%, 3.91\%] \\
\texttt{VLM (G)} & 226 & 6676 & 3.40\% & [2.98\%, 3.85\%] \\
\texttt{VLM (A)} & 219 & 6671 & 3.28\% & [2.87\%, 3.72\%] \\
\texttt{VLM (A+G)} & 222 & 6655 & 3.34\% & [2.93\%, 3.78\%] \\
\bottomrule
\end{tabular}}
\vspace{-1em}
\end{table}

The 95\%CrIs suggest that there is no strong evidence that any model meaningfully outperforms the others when compared to the ground truth (see Table \ref{cri}). The overlap of the intervals indicates that all models perform similarly, with only minor variations that remain within the range of uncertainty.


\begin{table}[h!]
\centering
\caption{Bayesian 95\% CrIs for Agreement Rate Differences. 
}\label{cri}
\begin{tabular}{lc}
\toprule
\textbf{Model Comparison} & \textbf{95\%CrI} \\
\midrule
\texttt{VLM} vs. \texttt{VLM (G)} & [-0.58\%, 2.02\%] \\
\texttt{VLM} vs. \texttt{VLM (A)} & [-1.69\%, 0.87\%] \\
\texttt{VLM} vs. \texttt{VLM (A+G)} & [-1.40\%, 1.17\%] \\
\texttt{VLM (G)} vs. \texttt{VLM (A)} & [-2.41\%, 0.16\%] \\
\texttt{VLM (G)} vs. \texttt{VLM (A+G)} & [-2.13\%, 0.45\%] \\
\texttt{VLM (A)} vs. \texttt{VLM (A+G)} & [-1.56\%, 0.88\%] \\
\bottomrule
\end{tabular}%
\vspace{-2em}
\end{table}

\subsection{Effect of Sensitive Attributes 
}

To evaluate if, beyond being consistent, the models are affected by the sensitive attributes of gender, we compared the assessment scores for boys and girls produced by \texttt{VLM (G)}  against the ground truth.  
For boys, there were 36 true positives (TP) and 103 false negatives (FN) ($n = 139$), yielding a TPR of 0.26, while girls had 34 TP and 79 FN ($n = 113$), resulting in a TPR of 0.30. The corresponding FPR for boys, calculated from 99 false positives (FP) and 2962 true negatives (TN) ($n = 3061$), was 0.03, compared to an FPR of 0.05 for girls (127 FP and 2573 TN; $n = 2700$). Table~\ref{tab:eo} summarizes the complete confusion matrix components and associated rates for both groups.

\begin{table}[h!]
\caption{ Error rate metrics by gender} 
\centering
\resizebox{\columnwidth}{!}{
\begin{tabular}{llcccccccc}
\toprule
\textbf{Method} & \textbf{Gender} & \textbf{TP} & \textbf{TN} & \textbf{FP} & \textbf{FN} & \textbf{TPR} & \textbf{FPR} & \textbf{TNR} & \textbf{FNR} \\
\midrule
\texttt{VLM (G)}  & Boys  & 36 & 2962 & 99  & 103 & 0.26 & 0.03 & 0.97 & 0.74 \\
\texttt{VLM (G)}  & Girls & 34 & 2573 & 127 & 79  & 0.30 & 0.05 & 0.95 & 0.70 \\
\midrule
\texttt{VLM}      & Boys  & 38 & 2985 & 105 & 102 & 0.27 & 0.03 & 0.97 & 0.73 \\
\texttt{VLM}      & Girls & 33 & 2603 & 125 & 85  & 0.28 & 0.05 & 0.95 & 0.72 \\
\bottomrule
\end{tabular}}
\label{tab:eo}
\end{table}

For the TPR, the pooled proportion across both groups was calculated as 0.28, \textit{SE} = 0.06, resulting in a z-value of 0.739. This difference (4.19 percentage points) was not statistically significant, $z(\infty)=0.739$, $p=0.46$, suggesting that the sensitivity of the model (\texttt{VLM (G)}) did not differ significantly between boys and girls. In contrast, the FPR showed a difference of 0.02 between groups. With a pooled negative proportion of 0.04, \textit{SE} = 0.01, the z-test produced a z-value of 2.87, which was statistically significant, $z(\infty)=2.87$, $p=.004$. This result indicates that the model (\texttt{VLM (G)}) produced a significantly higher false positive rate for girls than for boys. In summary, while the TPR differences across genders did not reach statistical significance, the significant disparity in FPR indicates that the model does not fully satisfy the EO criterion. Specifically, girls are more likely to be incorrectly assessed compared to boys, providing evidence of how sensitive attributes such as gender affect false positive predictions.

To further validate the source of the sensitive attributes, we compared the assessment scores for boys and girls produced by \texttt{VLM} against the ground truth. For boys, there were 38 TP and 102 FN ($n = 140$), yielding a TPR of 0.27, while girls had 33 TP and 85 FN ($n = 118$), resulting in a TPR of 0.28. The corresponding FPR for boys, calculated from 105 FP and 2985 TN ($n = 3090$), was 0.03, compared to an FPR of 0.05 for girls (125 FP and 2603 TN; $n = 2728$) (see Table~\ref{tab:eo}). 
For the TPR, the pooled proportion across both groups was calculated as 
 0.28, \textit{SE} = 0.06, resulting in a z-value of  0.15. This difference (0.82 percentage points) was not statistically significant, $z(\infty)=0.15$, $p = 0.88$, suggesting that the sensitivity of the model (\texttt{VLM}) did not differ significantly between boys and girls. In contrast, the FPR showed a difference of 0.01 between groups. With a pooled negative proportion of 0.04, \textit{SE} = 0.01, the z-test produced a z-value of = 2.31, which was statistically significant, $z(\infty)=2.31$, $p = 0.021$. This result indicates that the model (\texttt{VLM}) produced a significantly higher FPR for girls than for boys. In summary, while the sensitivity (TPR) of the model did not differ significantly between the two groups, the FPR was significantly higher for girls.

Spearman's correlation between the assessment of \texttt{VLM} and \texttt{VLM (A)} with participants' age confirmed that age was not a signficant sensitive attribute affecting the VLM's assessments ($\rho$ = 0.01, \textit{p} = 0.347; $\rho$ = -0.01, \textit{p} = 0.826). Sensitivity analysis is typically conducted for features with larger brackets~\cite{NIPS2016_9d268236}, so we decided not to further assess it.

\section{Discussion}


Our results highlight a critical discrepancy in assessment agreement and consistency. While negative assessments are detected with moderate reliability across all parsing methods, the agreement for actual assessments is alarmingly low, suggesting that model-generated outputs are far less effective in capturing positive assessments. This finding echoes previous findings (e.g.,~\cite{abbo2025vision,mehra2025beyond}), which highlight the limitations of VLMs and LLMs in classification tasks~\cite{Xu2024LLMsOverclaimed}.
This draws attention to the challenges of using VLMs and LLMs in tasks that demand a nuanced understanding of subjective content, where introduced to polarized values, or when it can benefit from human input (see \cite{dogan2024grace}). Future research can explore the sources of discrepancies in classification models, particularly concerning their sensitivity to positive assessments. 
Furthermore, 
different prompting strategies—whether incorporating age, gender, or both—yielded only marginal differences in assessment agreement and consistency. This suggests that demographic information alone does not substantially affect the model's ability to align with human assessments. More importantly, the VLM performed consistently regardless of whether it was provided with information about the child’s gender and age, indicating that these demographic factors did not significantly impact its overall performance. This consistency suggests that the model’s assessment mechanisms rely primarily on the content of the child’s narrative rather than demographic characteristics. In other words, it seems that the VLM's performance remains stable and is not sensitive to prompts including these demographic cues. 


Nevertheless, our findings also indicate that while the sensitivity of the VLM assessments (TPR) remains comparable across genders—with no significant differences—the models consistently yield a higher FPR for girls. This trend, observed in both the \texttt{VLM (G)} and the \texttt{VLM} assessments not prompted with gender information, suggests that the error distribution may still be inadvertently influenced by the sensitive attribute of gender, warranting careful reflection. This divergence, which contravenes the EO criterion, raises important questions regarding the underlying mechanisms that drive misclassification, especially in contexts where an erroneous positive prediction could have substantive implications. It is critical to acknowledge that while prompting the VLM with gender information did not affect the overall consistency of the VLM’s assessments, the elevated FPR for girls underscore a latent inequity that could impact the model's clinical decision-making. One possible explanation for the observed differences is that the models may be driven by the narratives shared by the participants, potentially affecting the model's performance. For instance, girls who are more likely to experience mental wellbeing concerns are more likely to share more expressive narratives than boys \cite{Chaplin2013GenderReview}, as previous research has found such differences in robot-led assessments~\cite{abbasi2024analysing}, as well as greater disclosure among individuals experiencing negative emotions~\cite{2023OpeningBehavior}. Future work should focus on probing the root causes of this disparity and on devising strategies to recalibrate the model, thereby ensuring that automated wellbeing assessments maintain both reliability and fairness across diverse populations.


Ethical concerns should also be taken into consideration before such technologies become more widespread. There are rising concerns with regard to over-reliance on technologies that might precipitate a decrease in human empathy in sensitive care settings~\cite{2024Sharing}. Thus, 
significant discussion from all relevant stakeholders is warranted to ensure necessary safeguards in supporting the wellbeing of children~\cite{levinson2024expert}. 


\section{Conclusions}
Our work has revealed compelling insights into the use of VLMs in the evaluation of children's wellbeing in a robot-led framework. While VLMs have shown moderate agreement with the psychologist's assessment in identifying negative classifications, their performance dropped considerably while evaluating narratives where assessments were present. Our findings also indicate that prompting the VLM with demographic information such as age and gender did not affect its overall consistency. Nonetheless, we found that the model's assessments were sensitive to gender, with girls more likely to be misclassified as having low wellbeing compared to boys. These findings underscore the need for careful scrutiny and bias mitigation when deploying VLMs in sensitive, child-focused contexts—especially when their outputs may inform real-world wellbeing decisions. 
Future research should explore the use of non-verbal cues (such as facial expressions and speech), additional demographic factors (including ethnicity and socioeconomic background), to develop a more holistic understanding of the effectiveness of VLM-powered robots in assessing children's wellbeing.


\bibliographystyle{IEEEtran}
\balance{\bibliography{ref}}

\end{document}